\documentclass[letterpaper, 10 pt, conference]{ieeeconf}  

\IEEEoverridecommandlockouts                              

\usepackage{graphicx}
\usepackage{multirow}
\usepackage{xcolor}
\usepackage{amsmath}
\usepackage[most]{tcolorbox}
\usepackage{hyperref}
\usepackage{booktabs}

\newcommand{\name}{\emph{SAFe-Copilot}}
\newcommand{\Intervention}{\mathcal{A}}

\newcommand{\state}{\mathcal{S}}
\newcommand{\lowlevelstate}{\state_{\text{low}}}
\newcommand{\highlevelstate}{\state_{\text{high}}}
\newcommand{\lowlevelobservation}{\mathcal{O}}
\newcommand{\environment}{\mathcal{E}}
\newcommand{\vlm}{\texttt{VLM}}
\newcommand{\policy}{\pi}
\newcommand{\autonomous}{\pi}

\newcommand{\plan}{\policy^{\ast}}
\newcommand{\lowlevelplan}{\tau_{\pi}}
\newcommand{\highlevelplan}{\hat{\tau}}
\newcommand{\trajectory}{\tau}
\newcommand{\planfunction}{\mathcal{G}}
\newcommand{\statefunction}{\mathcal{H}}
\newcommand{\LLMTextStyle}{\ttfamily\scriptsize}

\title{\LARGE \bf
\emph{SAFe-Copilot}: Unified \underline{S}hared \underline{A}utonomy \underline{F}ram\underline{e}work
}

\author{Phat Nguyen$^{1*}$, Erfan Aasi$^{1*}$, Shiva Sreeram$^{1}$, Guy Rosman$^{2}$, Andrew Silva$^{2}$, Sertac Karaman$^{1}$, Daniela Rus$^{1}$
\thanks{* These authors contributed equally.}
\thanks{- This work is supported by Toyota Research Institute (TRI). It, however, reflects solely the opinions and conclusions of its authors, and not TRI or any other Toyota entity.}
\thanks{$^{1}$MIT CSAIL, $^{2}$TRI, $^{3}$MIT LIDS}
}

\begin{document}

\maketitle
\thispagestyle{empty}
\pagestyle{empty}

\begin{abstract}


Autonomous driving systems remain brittle in rare, ambiguous, and out-of-distribution scenarios, where human driver succeed through contextual reasoning. Shared autonomy has emerged as a promising approach to mitigate such failures by incorporating human input when autonomy is uncertain. However, most existing methods restrict arbitration to low-level trajectories, which represent only geometric paths and therefore fail to preserve the underlying driving intent. We propose a unified shared autonomy framework that integrates human input and autonomous planners at a higher level of abstraction. Our method leverages Vision–Language Models (VLMs) to infer driver intent from multi-modal cues---such as driver actions and environmental context---and to synthesize coherent strategies that mediate between human and autonomous control. We first study the framework in a mock-human setting, where it achieves perfect recall alongside high accuracy and precision. A human-subject survey further shows strong alignment, with participants agreeing with arbitration outcomes in 92\% of cases. Finally, evaluation on the Bench2Drive benchmark demonstrates a substantial reduction in collision rate and improvement in overall performance compared to pure autonomy. Arbitration at the level of semantic, language-based representations emerges as a design principle for shared autonomy, enabling systems to exercise common-sense reasoning and maintain continuity with human intent.

\end{abstract}


\section{INTRODUCTION}
While autonomous driving approaches have sparked hopes for a transformed and better society \cite{milakis2017policy,bissell2020autonomous} and continue to grow in capability \cite{zhao2024autonomous}, their capabilities and coverage are still imperfect. Human factors also play a major role in traffic safety incidents \cite{mckenna1982human,hsiao2018preventing,bucsuhazy2020human}: although humans drivers excel at interpreting context, inferring intent, and applying flexible reasoning in edge cases, their judgment can be impaired by distraction~\cite{regan2011driver,foley2013towards,engstrom2017effects} and their responses are often slow~\cite{dozza2013factors,rosenholtz2017perceptual,drozdziel2020drivers}. Shared autonomy offers the potential to combine the strengths of both humans and autonomous systems, enabling safer driving~\cite{balachandran2021human,marcano2020review}.

Teaming the human driver and autonomy opens new possibilities, but it also introduces new challenges. Paradigms for human–AI teaming remain limited, constrained both by our ability to design them and by the AI’s lack of common sense about what effective teamwork on the road entails, including understanding the driver errors and their underlying causes. Moreover, aleatoric uncertainty about the driver's intent~\cite{fang2024behavioral} and situational awareness~\cite{biswas2024modeling,gopinath2021maad} further complicates fluent collaboration with the driver.

\begin{figure}[htpb]
    \centering
    \includegraphics[width=0.48\textwidth]{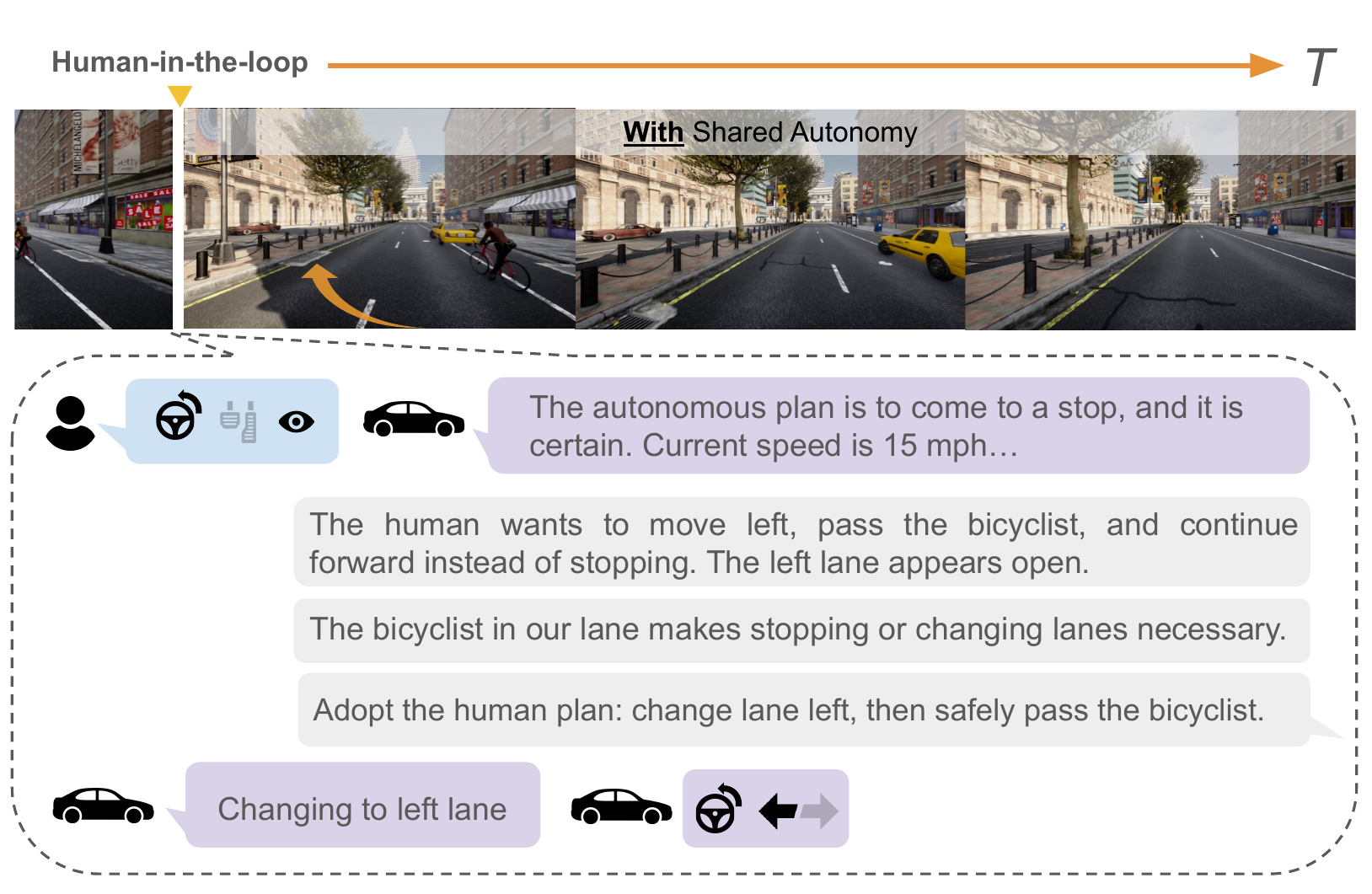}
    \caption{Given a driver intervention, \emph{\name} evaluates scene context, human intent, uncertainty, and autonomy's plans to arbitrate the most suitable and safest plan.}
    \label{fig:teaser}
    \vspace{-15pt}
\end{figure}

In this paper, we introduce \name, a unified shared autonomy framework that leverages the complementary strengths of human drivers and autonomous systems--ensuring safety while preserving the adaptability of human input (see Fig.~\ref{fig:teaser}).
Our approach uses Vision Language Model (VLM)-based policies to reason about and fuse human and autonomous plans in a synergistic way, exploiting the VLM’s commonsense reasoning to enable an elegant and generalizable form of shared autonomy. Specifically, our method infers high-level plans from human inputs (e.g., steering and pedal actions) and either arbitrates between the human and autonomy planners or blends elements of both, yielding more effective overall behavior of the driver–vehicle team~\cite{riener2012driver}.

More broadly, our framework treats shared autonomy as a principled process of combining human and machine plans into a coherent driving behavior. Given a driver action--such as steering or braking--together with the autonomous plan, the system arbitrates between them by either selecting one plan directly or merging their complementary aspects. The arbitration is guided by reasoning over safety, task objectives, and human intent, allowing the framework to integrate human flexibility without discarding the reliability of autonomy. In summary, our contributions are as follows:

\begin{itemize}
    \item {We introduce \name\, a shared autonomy framework that arbitrates between human and autonomous driving plans, unifying them into a consistent strategy that preserves the human intent.}
    \item {Our framework advances shared autonomy through: (i) integrating multimodal cues and environmental context into the arbitration process; (ii) merging human and autonomous plans at the level of abstract goals; and (iii) accommodating the flexible rule adherence often observed in human driving.}
    \item {We validate our approach through extensive experiments. In mock-human evaluations, the framework achieves perfect recall with high precision and accuracy. A human-subject survey further demonstrates strong alignment with human preferences, with participants agreeing with our arbitration framework in 92\% of cases, alongside significant reductions in collision rate and improvements in route completion compared to pure autonomy baselines.}
\end{itemize}



\section{RELATED WORKS}

Our approach is at the intersection of several fields of research. LLMs have recently found significant use in human-robot interactions \cite{zhang2023large,kim2024understanding}. Several avenues for their use~\cite{williams2024scarecrows} include language-based correction of control~\cite{cui2023no}, as well as general prompting of robotic policies~\cite{li2024towards} and code generation~\cite{wang2025automisty}.



In the field of autonomous and intelligent driving~\cite{yang2023llm4drive_survey}, language models are leveraged as scaffold for planners~\cite{LiLu2024DrivingWithInternVL,MTMM2024MMAD} and policies~\cite{wang2024drive, xu2024drivegpt4,sima2024drivelm,ma2024lampilot}, as well as predictors~\cite{kuo2022trajectory,seff2023motionlm}. Other uses include interface approaches for speech-guided driving~\cite{deruyttere2019talk2car,cui2024personalized,driveasyouspeak2023}, or as approaches for augmented data generation and policy training~\cite{zhong2023language,tan2024promptable,aasi2025generating,nguyenregen}. This is in addition to holistic models trained on large-scale data sources~\cite{wayve2023lingo1,hwang2024emma}.

Finally, shared control and shared autonomy in general have been heavily explored in the context of semi-autonomous driving~\cite{marcano2020review,xing2020driver,wang2020decision}. Several works have both characterize possible approaches~\cite{abbink2018topology} and address their different limitations~\cite{de2023shared}.
Approaches for understanding how to share control in the space of high-level plans have drawn recent attention, with both classical control~\cite{weiss2023follow} and reinforcement-learning based~\cite{fisac2019hierarchical,decastrodreaming} approaches.


\section{TECHNICAL APPROACH}
\label{sec:method}

\begin{figure*}[t]
    \vspace{5pt}
    \centering
    \includegraphics[width=0.99\textwidth]{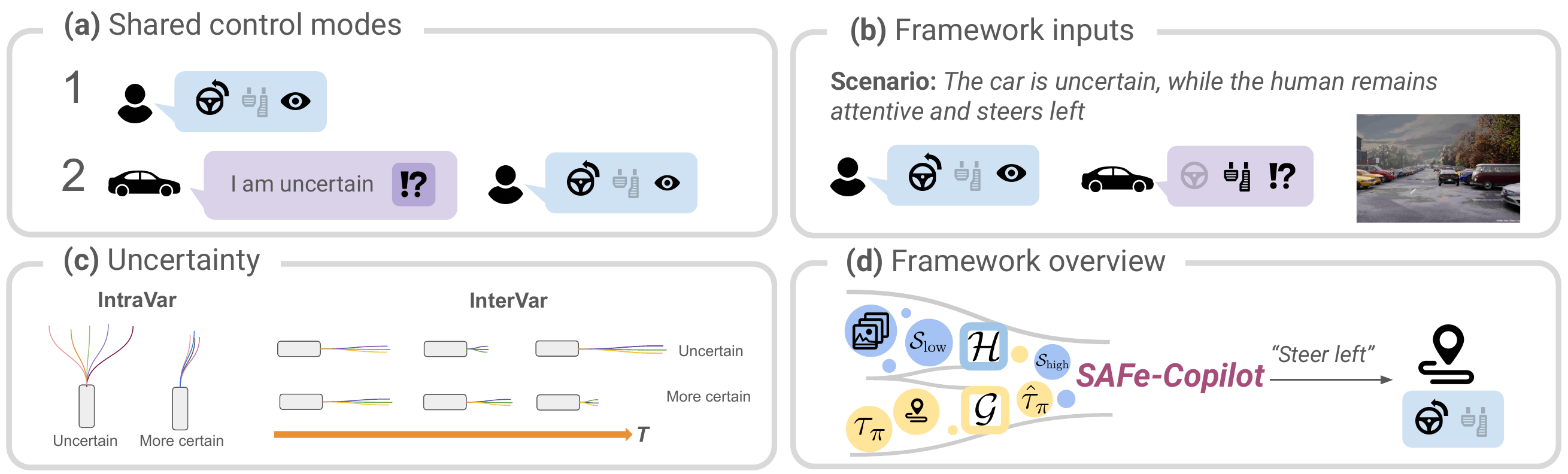}
    \caption{\textbf{Overview.} \emph{(a)} Our framework supports two teaming modes: \emph{Top}: proactive teaming that fuses driver input with autonomy whenever the human intervenes; \emph{Bottom}: supervisory shared control where the system requests human input under high uncertainty. \emph{(b)} The system takes as input the driver’s state and control actions, along with the ego-vehicle state. \emph{(c)} Intra-frame variance measures disagreement among candidate trajectories in a single frame, while inter-frame variance measures changes in the mean trajectory across frames. \emph{(d)} \name\ integrates driver state and actions, vehicle state, and uncertainty measures within a symbolic reasoning module that arbitrates between human and autonomous plans to generate a coherent and safe trajectory.}
    \label{fig:method}
\end{figure*}

 \name\  is a unified framework for fusing predicted human and autonomous system's plans in interactive driving scenarios (see Fig.~\ref{fig:method}). The framework takes as input human interventions $\Intervention$---such as steering, throttle, or braking actions---along with a representation of the environment $\environment$, and outputs a trajectory $\trajectory \sim \plan$ generated under the plan $\plan$ selected by our arbitration process. The environment is represented as $\environment = (\lowlevelobservation, \lowlevelstate)$, where $\lowlevelobservation$ corresponds to perceptual inputs (e.g., image from the ego vehicle's forward-facing camera), and $\lowlevelstate$ consists of the human state (e.g., attention or gaze direction \cite{gopinath2021maad}) together with the ego-vehicle state (e.g., position, velocity, and heading).



Our framework enables several modes for teaming---specifically, we examine in this paper two complementary modes: (i) \emph{proactive teaming}, where system arbitrates and fuses between the driver controls and suggested autonomy plans whenever the human intervenes; and (ii) \emph{prompted, supervisory, shared control ~\cite{sheridan2012human}}, where the system explicitly requests human input when an uncertainty detector flags low confidence in the autonomy plan. In both cases, human actions are incorporated directly into the arbitration process, ensuring that they influence not just the immediate vehicle control but also the selection of the future trajectory. 
This design provides drivers with various degrees of agency, maintaining high recall, precision, and accuracy in trajectory selection.


Formally, we express $\trajectory \sim \texttt{\name}(\Intervention, \environment)$, where \name\ consists of three components: (i) \textit{Abstraction module}: a translator that maps raw state variables into high-level descriptions (Sec.~\ref{ssec:lowlevel}); (ii) \textit{Uncertainty module}: detects conditions in which the autonomous plan is unreliable and determines when human intervention is required (Sec.~\ref{ssec:uncertainty}); and (iii) \textit{Reasoning module}: a video–language reasoning module that leverages VLMs to integrate contextual information into the arbitration between human and autonomous plans (Sec.~\ref{ssec:reasoning}).

\subsection{Abstraction Module}
\label{ssec:lowlevel}

In this section, we describe how continuous, low-level signals are abstracted into discrete, high-level descriptors. Given a representation of the environment $\environment = (\lowlevelobservation, \lowlevelstate)$ and the low-level plan $\tau_{\policy} = \{(x_i, y_i)\}_{i=0}^{T}$ from the autonomous planner $\policy$, with each $(x, y)$ denoting a waypoint in Cartesian coordinates, the abstraction module produces a high-level plan $\highlevelplan_{\policy} = \planfunction(\lowlevelplan)$ and a high-level state $\highlevelstate = \statefunction(\lowlevelstate)$.

\subsubsection{Plan Abstraction $\planfunction$}
A low-level plan $\lowlevelplan$ is summarized by its net lateral and longitudinal displacement, as well as its total path length:
\begin{equation}
    \begin{split}
        & \Delta x = x_T - x_0, \ \ \  \Delta y = y_T - y_0,  \\
        & L = \sum_{i=1}^{T} \sqrt{(x_i - x_{i-1})^2 + (y_i - y_{i-1})^2}.
    \end{split}
\end{equation}
Here, the displacement $(\Delta x, \Delta y)$ represents the overall direction of motion, while $L$ measures the trajectory length. Using thresholds $\theta_{\text{stop}}, \theta_{\text{turn}}, \theta_{\text{fwd}}$, the plan is categorized into discrete high-level behaviors:
\begin{itemize}
    \item If $L < \theta_{\text{stop}}$, the plan is labeled as stop;
    \item If $|\Delta x| > \theta_{\text{turn}}$, the plan is labeled as a ``turn'', with the sign of $\Delta x$ distinguishing left from right;
    \item If $\Delta y > \theta_{\text{fwd}}$, the plan is labeled as ``drive forward'', whereas smaller forward progress is labeled as ``slow down'';
\end{itemize}

\subsubsection{State Abstraction $\statefunction$}
The low-level state $\lowlevelstate$ is discretized into textual descriptors. Throttle and brake values are binned into intervals $B=\{0.0,0.25,0.5,0.75,1.0\}$ and mapped to ordinal labels
\[
\mathcal{L} = \{\text{not applied}, \text{light}, \text{moderate}, \text{strong}, \text{maximum}\}.
\]
Steering values are classified using an angular threshold $\epsilon_\theta$: values greater than $\epsilon_\theta$ are labeled ``to the right'', values less than $-\epsilon_\theta$ as ``to the left'', and values in between as ``neutral''. Speed values are converted into miles per hours. The human state abstraction is defined heuristically in this setup, though more principled methods, such as synthesizing the driver's attended awareness \cite{gopinath2021maad}, could also be employed.

\subsubsection{Practical Considerations}
In this paper, We set $\theta_{\text{stop}}=1.5$, $\theta_{\text{turn}}=2.0$, $\theta_{\text{fwd}}=2.0$, and $\epsilon_\theta=0.05$ rad. These thresholds were chosen empirically and yield stable abstractions across diverse driving routes. The autonomous plans are processed through $\planfunction$, while the ego and human control states are abstracted through $\statefunction$.





\subsection{Uncertainty Module} 
\label{ssec:uncertainty}

We incorporate an uncertainty module that estimates when the autonomous policy $\autonomous$ produces unreliable plan, favoring human intervention in such cases. The module takes as input a sequence of candidate trajectories predicted by $\autonomous$ across consecutive frames and outputs an uncertainty score $u_t \in [0,1]$ indicating the system’s confidence in executing the autonomous plan.

The uncertainty is estimated from two complementary sources of variability. The first is \emph{intra-frame variance} $\text{IntraVar}_t$, which measures how much the set of candidate trajectories differ within a single frame. A high intra-frame variance indicates that the policy is indecisive about its immediate motion. The second is \emph{inter-frame variance} $\text{InterVar}_t$, which measures how much the mean trajectory changes across consecutive frames; high inter-frame variance indicates temporal inconsistency in the policy’s predictions.

These two measures are combined into a single uncertainty score as a convex combination:
\begin{equation}
    u_t = \frac{\alpha \cdot \text{IntraVar}_t + \beta \cdot \text{InterVar}_t}{\alpha + \beta},
\end{equation}
where $\alpha$ and $\beta$ are hyperparameters that balance the two contributions. The resulting score $u_t$ indicates when the autonomous planner is uncertain, with larger values of $u_t$ corresponding to greater planner uncertainty. Note that alternative uncertainty estimates could also be incorporated here, such as planner- or subsystem-level confidence measures~\cite{buhler2020driving}, cost- or value-based metrics~\cite{fisac2018probabilistically}, or scenario-level out-of-distribution detectors~\cite{aasi2025generating}.


\subsubsection{Practical Considerations}
Both $\text{IntraVar}_t$ and $\text{InterVar}_t$ are normalized to ensure comparability across scenarios, and the weights $\alpha$ and $\beta$ are determined empirically. A threshold $\theta_u$ is then applied, and human intervention is requested at the first frame $t$ where $u_t > \theta_u$. This design allows the uncertainty module to act as a safeguard, triggering human input whenever the autonomous system lacks confidence in its own plan.

\subsection{Reasoning Module}
\label{ssec:reasoning}

The reasoning module takes as input a high-level textual description $\highlevelstate$ together with a sequence of RGB images from the ego-vehicle's forward-facing camera, $\mathbf{o}_{t-2:t} = (o_{t-2}, o_{t-1}, o_t)$. These 3 frames are sampled at $0.5$-second intervals over a $1$-second horizon, consistent with prior works \cite{sreeram2024probingmultimodalllmsworld}, \cite{hwang2024emmaendtoendmultimodalmodel}. The module outputs a selected plan $\plan$, which is then used to generate a trajectory $\trajectory$. Formally, we define the reasoning function as $\plan \sim \vlm(\highlevelstate, \mathbf{o}_{t-2:t})$,
where $\vlm$ denotes the VLM-based reasoning module implemented with ChatGPT \texttt{o3-2025-04-16}~\cite{openai_o3_system_card_2025}. In principle, $\plan$ may correspond to a learned policy \cite{nguyen2024texttodrivediversedrivingbehavior} or to a library of motion primitives. In this work, we adopt motion primitives for their structured representation and compatibility with standard planning and control frameworks.


Specifically, VLM is used for: (i) analyzing visual inputs by leveraging VLM's inherit object classification and scene understanding to extract semantic description of the environment, (ii) inferring human intent from multi-modal cues---such as steering and pedal inputs---together with the visual context to infer human intent, and (iii) grounding arbitration decisions in symbolic abstractions, which enables the system to either choose between the autonomous plan and the human plan, or synthesize a fused plan that merges aspects of both. This structured use of the VLM provides a principled mechanism for integrating human and autonomous driving plans. A shortened version of the prompt is as below:

\begin{tcolorbox}[colback=gray!10, colframe=gray!50!black, boxrule=0.5pt, arc=2mm]
\LLMTextStyle PROMPT: Given the image, your task is to analyze the situation and describe what is happening in the scene. Then you are given a human intervention query that describes the human's actions. Your task is to infer the plausible intentions of the human driver when such interventions occur. Based on your inference, you are tasked with either choosing the human plan, the autonomous plan, or propose an alternative.

Ego-vehicle state: Throttle: ..., Steering: [...]

Autonomous stack plan: [...]. Human state: [...]

...
\end{tcolorbox} 

\section{RESULTS}
\begin{figure*}[t]
    \vspace{5pt}
    \centering
    \includegraphics[width=0.96\textwidth]{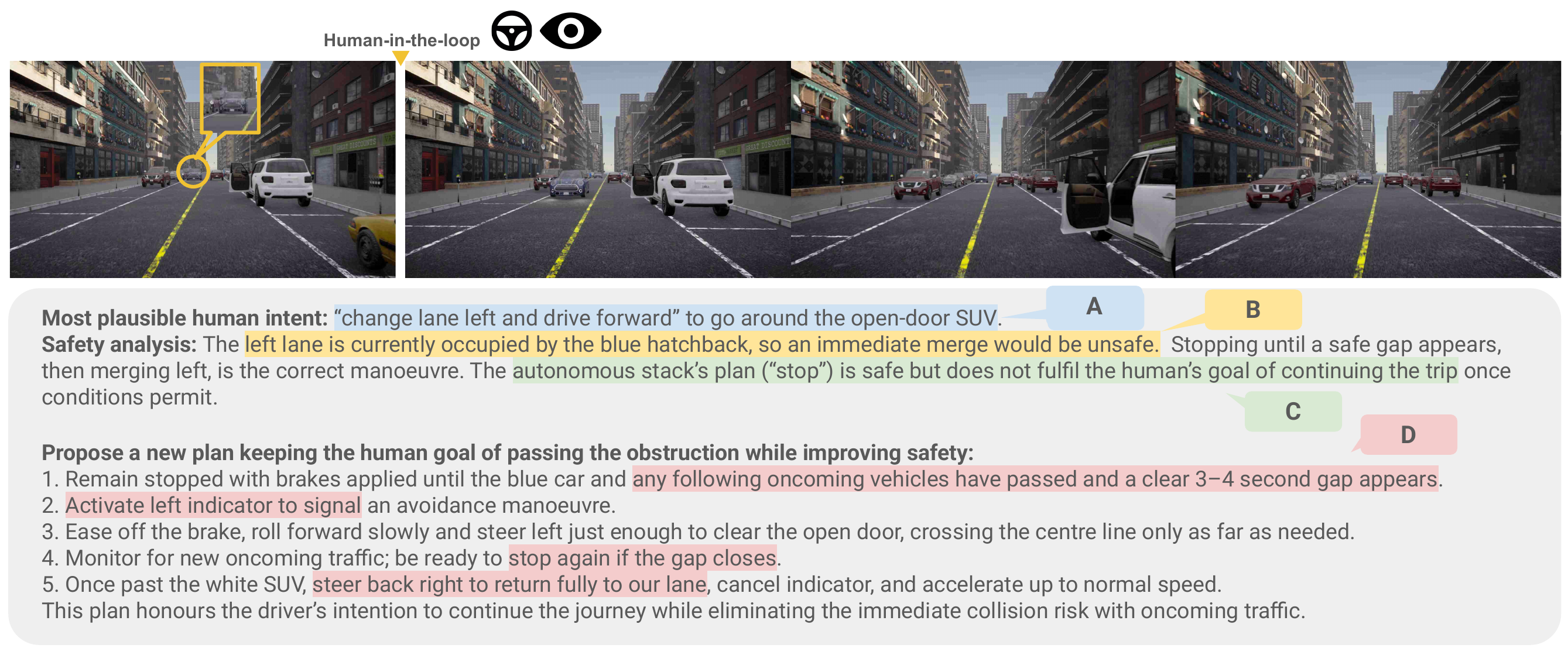}

    \caption{\textbf{Qualitative example.} Output for a scenario in which the driver steers left to avoid an open car door obstructing the lane, while oncoming traffic approaches. The results show that the framework: \emph{(A)} correctly infers human intent, \emph{(B)} evaluates the consequences of the human plan, \emph{(C)} contrasts it with the autonomous plan, and \emph{(D)}  demonstrates an understanding of societal driving norms, leveraging them to fuse both plans into a safer trajectory. }
    \label{fig:qualitative}
    \vspace{-10pt}
\end{figure*}

\begin{figure}[t]
    \centering
    \includegraphics[width=0.48\textwidth]{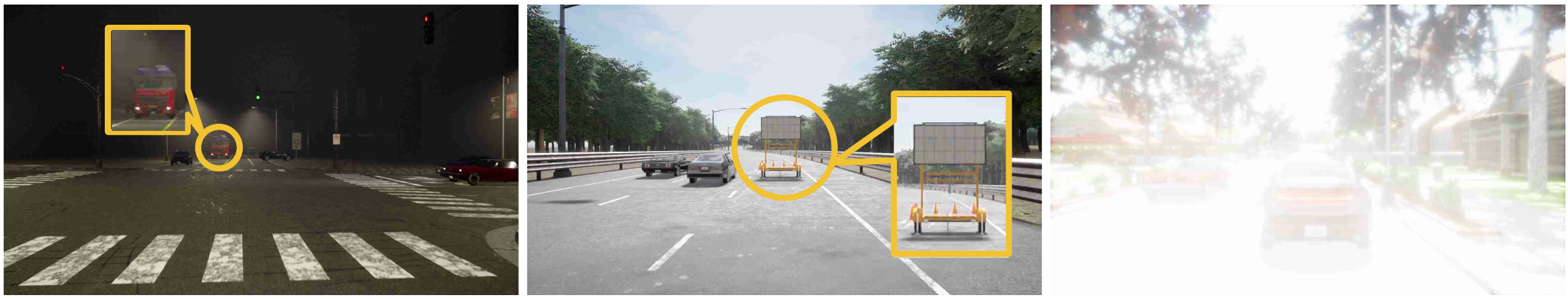}
    \caption{\textbf{Scenario examples.} \emph{Left}: yielding to emergency vehicle. \emph{Middle}: overtaking an OOD construction sign. \emph{Right}: Glare perception failure}
    \label{fig:qualitative2}
    \vspace{-10pt}
\end{figure}



In this section, we evaluate our framework under the two operating modes of shared autonomy. The first mode implements full teaming between autonomous and human plans based on uncertainty and confidence cues. To study this setting, we construct a set of experimental scenarios, with qualitative results shown in Sec.~\ref{ssec:qualitative}. Sec.~\ref{ssec:mock_human} further evaluates this mode using an intentionally imperfect mock human, demonstrating our frameworks perfect recall alongside high accuracy and precision. A human survey (Sec.~\ref{ssec:human_survey}) shows strong alignment with our framework, with participants agreeing with the arbitration outcomes in 92\% of cases. The second mode follows the traditional paradigm of supervisory shared control~\cite{sheridan2012human}. We evaluate this mode on the Bench2Drive benchmark, observing a  reduction in collision rate and a significant improvement in driving score (Sec.~\ref{ssec:b2d}). 



\subsection{Experimental Setup} 
\label{ssec:experimental_setup}
In our shared autonomy setting, the \emph{ego} vehicle is equipped with both an autonomous policy and a human driver, with overall control mediated by our framework. Experiments are conducted in the CARLA simulator~\cite{dosovitskiy2017carla}, using the Bench2DriveZoo~\cite{jia2024bench2drive} setup. Autonomous policy plans are produced by VAD~\cite{jiang2023vad}, while human plans are specified as part of the scenario design and categorized as either \emph{correct} or \emph{incorrect}. A plan is considered correct if it avoids collisions and achieves route completion. We curated a total of 40 scenarios under this setup, which are further evaluated through a human study described in Sec.~\ref{ssec:human_survey}. The scenarios include overtaking around an opening car door across solid yellow lines, yielding to emergency vehicles, changing lanes for a highway closure, and handling abrupt maneuvers from a distracted driver. For each scenario, our framework arbitrates between the autonomous and human plans. If the autonomous plan is selected, the vehicle proceeds under pure autonomy. If the human plan is selected, the framework instantiates the corresponding motion primitives (e.g., driving forward, decelerating, or changing lanes). Path planning is performed using A$^\ast$ search, and low-level control is implemented via PID controllers.

\subsection{Qualitative Analysis}
\label{ssec:qualitative}

Here we present qualitative results that highlight the capabilities of our framework. For each scenario, the framework analyzes the scene given the image, the human action, and the autonomous plan, and produces an inferred intention and a plan decision. We illustrate representative cases that demonstrate the distinct capabilities of our framework.

\textbf{Multi-modal intention prediction.} The framework infers driver intent by integrating perceptual inputs with behavioral cues into a semantic description of the intention. In the example below, object-level recognition from the visual input is combined with cues such as braking intensity and gaze toward cross-traffic, as shown in Fig.~\ref{fig:qualitative2} (left). The resulting textual abstraction, shown below, expresses the driver’s likely intent (yielding to the emergency vehicle) which aligns with the situational context and informs downstream decision-making.

\begin{tcolorbox}[colback=gray!10, colframe=gray!50!black, boxrule=0.5pt, arc=2mm]
\LLMTextStyle Output: A large red fire truck with emergency lights is coming toward us in the opposite lane and will shortly enter the intersection. [...] Given the hard braking and the fact that the driver glanced to the left, the most plausible intention is to bring the car to a stop in order to yield—most likely to the approaching emergency vehicle or to some cross-traffic they may have noticed. [...] Continuing straight at 20 mph while an emergency vehicle approaches could create risk or violate yield laws.
\end{tcolorbox} 

\textbf{Common-sense reasoning.} The system leverages the VLM’s common-sense reasoning not only to contextualize the driver’s intent but also to exercise foresight at a higher level of abstraction. In this case, it evaluates the risks of stopping on the highway and identifies the associated trade-offs, in the scenario seen in Fig.~\ref{fig:qualitative2} (middle). This enable the system to balance the human and autonomous plans while maintaining strategic alignment with long-term driving goals in the shared autonomy setting.

\begin{tcolorbox}[colback=gray!10, colframe=gray!50!black, boxrule=0.5pt, arc=2mm]
\LLMTextStyle Output: [...] Stopping in the right lane will avoid an immediate collision, but it will leave the car stationary in a closed lane, forcing a later merge from a stand-still and possibly blocking following traffic.  
\end{tcolorbox} 

\textbf{Symbolic plan merging.} Fig.~\ref{fig:qualitative} shows a case where neither the driver's plan nor the autonomous plan is individually sufficient: the driver’s intent may conflict with safety requirements, while the autonomous plan may sacrifice progress for caution. At the trajectory level, such plans cannot be directly combined---interpolating between low-level motions strips away their semantic meaning and produces trajectories that no longer correspond to any coherent driving intent. To resolve this, the reasoning module operates at a symbolic abstraction layer, where plans are presented in terms of higher-level goals. At this level, the system merges elements of both plans into a composite strategy that preserves continuity of intent. 

\textbf{Relation to Safety Contracts.} An emergent property of our framework is the capacity to recognize formal traffic rules while simultaneously accommodating the pragmatic flexibility exhibited by human drivers. In the example of Fig.~\ref{fig:qualitative}, crossing a yellow solid line is treated not as an outright violation but as a temporary maneuver to maintain progress, with the implicit expectation of merging back once safe. Our framework arbitrates this as a contextualized ``contract," capturing the notion that human-like driving often requires controlled deviations from strict rules in order to maintain overall safety and traffic flow. This highlights how shared autonomy can evolve beyond rigid rule-following to accommodate flexible human driving styles, which lend itself to LLM-based safety reasoning~\cite{manas2024tr2mtl}.

\textbf{Sensor failure.} Our framework remains robust under sensor degradation. In scenarios with severe sun glare that washes out the camera input, as in Fig.~\ref{fig:qualitative2} (right), the reasoning module detects the failure and defers to human input as part of the collaborative driving process.

\begin{tcolorbox}[colback=gray!10, colframe=gray!50!black, boxrule=0.5pt, arc=2mm]
\LLMTextStyle Output: The camera view is almost completely washed-out by intense sun-glare/over-exposure. [...] plausible intent is to bring the car to a stop to avoid a potential collision with the barely visible vehicle ahead. Continuing to accelerate when visibility is severely compromised is unsafe.
\end{tcolorbox}

\subsection{Mock Human}
\label{ssec:mock_human}

\begin{table*}[t]
\vspace{5pt}
    \resizebox{\textwidth}{!}{%
\begin{tabular}{c|ccc|ccc|ccc|ccc}
\hline
\multirow{2}{*}{\textbf{Mock Human}} & \multicolumn{3}{c|}{\textbf{Accuracy (\%) $\uparrow$}}                                 & \multicolumn{3}{c|}{\textbf{Precision (\%) $\uparrow$}}                                     & \multicolumn{3}{c|}{\textbf{Recall (\%) $\uparrow$}}                                        & \multicolumn{3}{c}{\textbf{F1 (\%) $\uparrow$}}                                             \\ \cline{2-13} 
                                     & \multicolumn{1}{c|}{Naive} & \multicolumn{1}{c|}{D. Tree} & Ours            & \multicolumn{1}{c|}{Naive} & \multicolumn{1}{c|}{D. Tree} & Ours            & \multicolumn{1}{c|}{Naive} & \multicolumn{1}{c|}{D. Tree} & Ours            & \multicolumn{1}{c|}{Naive} & \multicolumn{1}{c|}{D. Tree} & Ours            \\ \hline
75\%                                 & 75.00                      & 58.62                        & \textbf{100.00} & 75.00                      & 61.54                        & \textbf{100.00} & 100.00                     & 53.33                        & \textbf{100.00} & 85.71                      & 57.14                        & \textbf{100.00} \\
50\%                                 & 50.00                      & 58.62                        & \textbf{93.33}  & 50.00                      & 61.54                        & \textbf{88.24}  & 100.00                     & 53.33                        & \textbf{100.00} & 66.67                      & 57.44                        & \textbf{93.75}  \\
25\%                                 & 25.00                      & 62.07                        & \textbf{90.00}  & 25.00                      & 64.29                        & \textbf{80.00}  & 100.00                     & 60.00                        & \textbf{100.00} & 40.00                      & 62.07                        & \textbf{88.89}  \\ \hline
\end{tabular}
    }%

\caption{\textbf{Mock human.} Evaluation of arbitration under varying human reliability}
\label{tab:mock_human}
\vspace{-10pt}
\end{table*}

\begin{figure*}[t]
    \centering
    \includegraphics[width=0.9\textwidth]{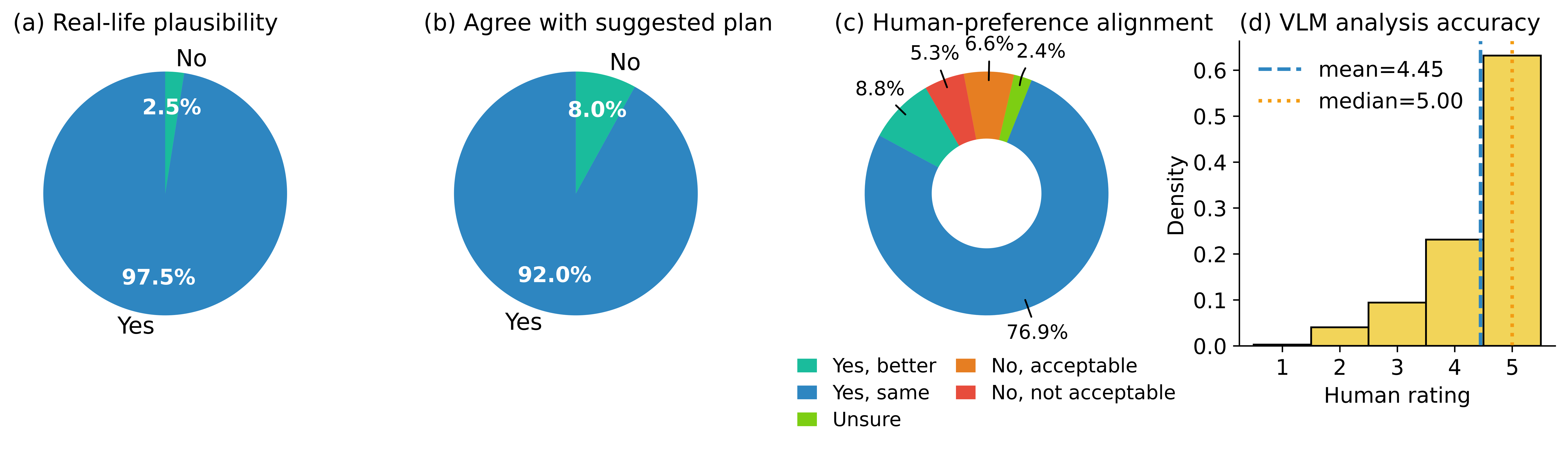}

    \caption{\textbf{Human Survey.} Human survey over sample scenarios showing \emph{(a)} overwhelming majority found the scenarios plausible; \emph{(b)} a strong majority agreed with the plan suggested by the VLM; \emph{(c)} Most participants agreed with the arbitration outcome, and a substantial minority judged it an improvement over their own plan; and \emph{(d)} overall annotators found the VLM’s analysis of the situations---such as predicting human intent---to be highly accurate.}
    \label{fig:human_survey}
    \vspace{-10pt}
\end{figure*}

\autoref{tab:mock_human} evaluates our method against two baselines---Naive and Decision tree \cite{IOVINO2022104096}, \cite{9827035}---using a mock human that proposes the correct plan in 75\%, 50\%, and 25\% of cases. A plan is considered correct if it avoids immediate collisions and allows the vehicle to complete its route. The naive baseline corresponds to always selecting the human plan, while the decision tree baseline is constructed by first prompting a LLM to generate a set of rule-based conditions, and then prompting another LLM to apply this decision tree to arbitrate between the human and autonomous plan. 

Our method achieves \emph{perfect recall across all reliability levels}, matching the naive baseline. This means that whenever the human proposes a correct plan, our system consistently selects it. Unlike the naive baseline, however, our approach maintains substantially higher accuracy and precision, demonstrating its ability to reject incorrect human plans as well. In this way, the framework \emph{acts as a guardian and can arbitrate effectively when the human is imperfect}.

In comparison to the decision tree baseline, our framework exhibits greater flexibility by leveraging the contextual reasoning capabilities of VLMs and avoiding over-reliance on rigid rule structures. Decision trees enforce deterministic constraints---for instance, categorically disallowing maneuvers across solid yellow lines---which can incorrectly exclude contextually safe actions and result in unwarranted failures. Our method, in contrast, combines symbolic reasoning with contextual evaluation to arbitrate adaptively. This allows the system to uphold correct human inputs while overriding unsafe or unnecessarily restrictive proposals, thereby establishing a robust “safety contract” between the human driver and the autonomous system.

\subsection{Human Survey}
\label{ssec:human_survey}

To capture public perceptions of our shared‑autonomy framework, we developed an online questionnaire featuring 40 driving scenarios described in Sec.~\ref{ssec:experimental_setup}. For each scenario, participants were presented with the same inputs as the VLM---three frames sampled at 0.5-s intervals, a textual description of the environment, and key states such as vehicle speed and driver attentiveness. For each scenario, participants were first asked whether the situation could plausibly occur in real life (Fig.~\ref{fig:human_survey}(a)), and then whether they agreed with the plan suggested by our shared-autonomy framework (Fig.~\ref{fig:human_survey}(b)). To further probe alignment with participant preferences, we asked them to act as the arbitrator between the human driver and the autonomous system. The response options were: “Yes---better than what I had in mind” (“Yes, better”), “Yes---I would have proposed the same” (“Yes, same”), “No---I have a better plan, but this one is acceptable” (“No, acceptable”), “No---I have a better plan, and this one is not acceptable” (“No, not acceptable”), and “I would not know what to do here” (“Unsure”). Finally, participants rated the VLM’s accuracy in analyzing the scene and explaining its decisions on a 1-5 scale (Fig.~\ref{fig:human_survey}(d)). We collected 38 responses, with an average participant age of 26.5 years (SD = 4.8) and average driving experience of 3.3 years (SD = 2.2). Of the respondents, 14 reported prior use of a VLM and 24 had rode in fully autonomous vehicles.

As shown in Fig.~\ref{fig:human_survey}(b), most participants judged the scenarios as plausible and agreed with the arbitration outcome (92$\%$). When asked how they would arbitrate if acting as the arbitrator, $76.9\%$ selected “Yes, same”, and a substantial minority judged it to be an improvement over their own plan (8.8$\%$ “Yes, better”), yielding $\approx 85.7\%$ direct alignment. Only 6.6$\%$ viewed our plan as acceptable but inferior, 5.3$\%$ found it unacceptable, and 2.4$\%$ were unsure. Respondents also largely agreed the scenarios as plausible and expressed high agreement with our proposed plans. In addition, the VLM’s scene analysis and explanations were rated highly. Taken together, these results suggest that our arbitration closely matches human preferences across the tested situations and support SAFe-Copilot as an effective mediator between human and autonomous policies, achieving a blend that participants report as matching or surpassing their own plans in most cases. While the study is modest in size and based on simulator scenarios, the evidence points to a language-driven shared-autonomy framework that people both agree with and trust.

\begin{table}[h]
    \resizebox{\columnwidth}{!}{%
        \begin{tabular}{c|cccc}
        \hline
        \multirow{2}{*}{\textbf{Models}} & \multicolumn{4}{c}{\textbf{Models}}                                                                                           \\ \cline{2-5} 
                                         & \multicolumn{1}{l}{UniAD-Tiny} & \multicolumn{1}{l}{UniAD-Base \cite{hu2023planningorientedautonomousdriving}} & \multicolumn{1}{r|}{VAD \cite{jiang2023vad}}   & \multicolumn{1}{r}{VAD + Ours} \\ \hline
        Collision Rate $\downarrow$      & 48.37                          & 51.37                          & \multicolumn{1}{c|}{46.11} & \textbf{38.89}                 \\
        Route Completion Rate $\uparrow$ & 63.49                          & 68.17                          & \multicolumn{1}{c|}{65.53} & \textbf{74.19}                 \\
        Average Score $\uparrow$         & 41.68                          & 46.23                          & \multicolumn{1}{c|}{43.74} & \textbf{55.97}                 \\ \hline
        \end{tabular}%
    }
    \caption{\emph{Bench2Drive Evaluation.} Performance of shared autonomy compared to pure autonomy.}
    \label{tab:bench2driveeval}
    \vspace{-20pt}
\end{table}

\begin{figure*}[t]
    \vspace{5pt}
    \centering
    \includegraphics[width=0.9\textwidth]{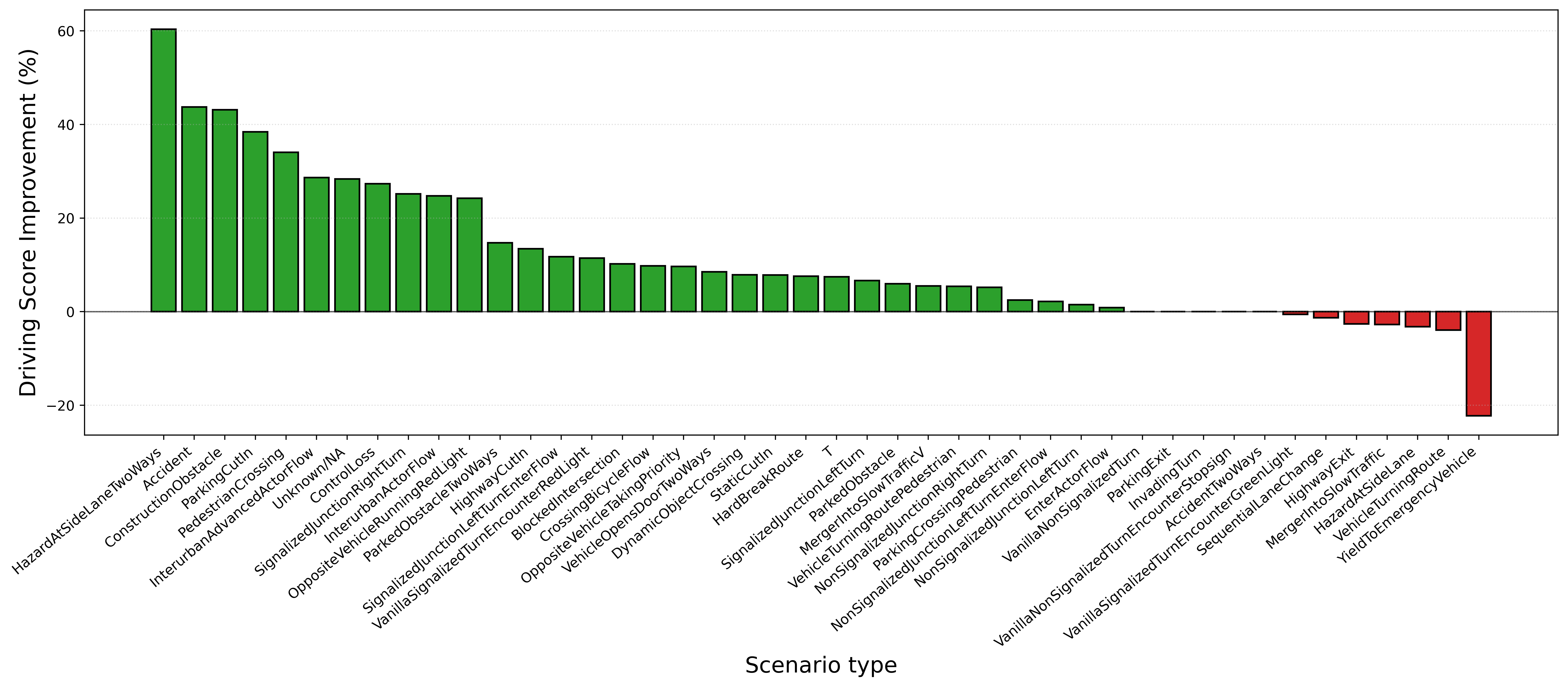}
    \caption{\textbf{Driving score improvements.} Scenario-wise improvements of shared autonomy over pure autonomy, evaluated on the Bench2Drive benchmark using a composite score that combines collision rate and route completion.}
    \label{fig:b2d_improvement_score}
    \vspace{-10pt}
\end{figure*}

\begin{figure}[h]
    \centering
    \begin{tabular}{cc}
        \includegraphics[width=0.9\columnwidth]{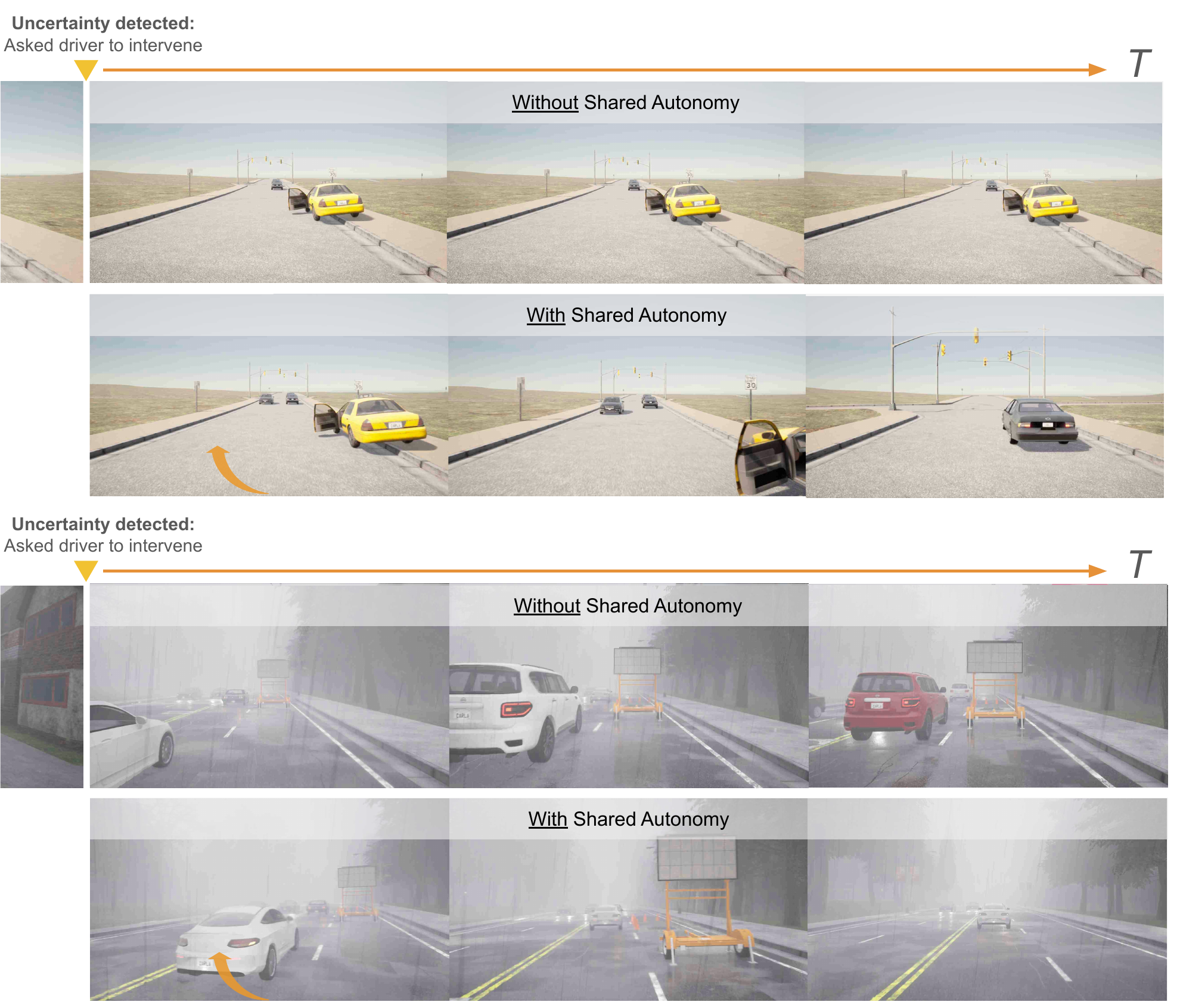} &
    \end{tabular}
    \caption{\textbf{Qualitative examples.} Results on the Bench2Drive benchmark.}
    \label{fig:b2d_qualitative}
    \vspace{-10pt}
\end{figure}

\subsection{Bench2Drive Evaluation} 
\label{ssec:b2d}
\autoref{tab:bench2driveeval} presents results on the Bench2Drive benchmark~\cite{jia2024bench2drive}, comparing our shared autonomy framework, with the VAD policy \cite{jiang2023vad}, against other baselines~\cite{hu2023planningorientedautonomousdriving}. Evaluated across 180 scenarios, the framework reduces collision rate by $15.66\%$, increases route completion rate by $13.22\%$, and improves the route composite score---which combines collision and route completion---by $27.96\%$. These results demonstrates that our shared autonomy framework enhances the policy performance, achieving superior outcomes compared to pure autonomy.

Fig.~\ref{fig:b2d_improvement_score} presents a breakdown of performance gains across different scenario types. We observe that many improvements arise in Out-Of-Distribution (OOD) situations, such as construction zones, opening car doors, or partially parked vehicles obstructing the lane. With a recall of 90\%, the uncertainty module identifies frames where the autonomous policy exhibits elevated uncertainty and prompts human input. Without arbitration, the base policy (e.g., VAD) frequently becomes indecisive and halts, or attempts an overtake that leads to collision, consistent with observations from Bench2DriveZoo~\cite{li2024think}. Representative examples are provided in Fig.~\ref{fig:b2d_qualitative}. For example, in the bottom row of Fig.~\ref{fig:b2d_qualitative}, the base policy encounters an out-of-distribution object under hazardous rainy conditions. The autonomous plan is to stop, but this leads the vehicle to remain too close and ultimately collide when proceeding forward. With human intervention, however, our framework selects a lane change and overtake as the safer alternative, thereby avoiding collision and completing the route. Other failure cases include lane invasions and collisions with neighboring vehicles. In such cases, the human corrects abrupt steering (e.g., a sharp jerk to the left), and the VLM infers the intention as ``re-centering in the lane while yielding to the adjacent vehicle before merging.''

\section{CONCLUSION}
In this paper we presented \emph{SAFe-Copilot}, a framework for VLM-based sharing of autonomy between driver and AI plans. Our approach demonstrates versatility in arbitrating and fusing plans at both the planning and the control level, easily accomodates reasoning of uncertainty and confidence, and leverages the common sense reasoning of LLMs towards better reasoning over driving rare events.

\textbf{Limitations}
We acknowledge that more extensive online human-in-the-loop experimentation, or self-verification of resulting plans are beyond the scope of the current work. They form a basis for future avenues of research.







\bibliographystyle{plain}
\bibliography{source/references}

\end{document}